%% file: main.tex
\documentclass{article}

\usepackage{microtype}
\usepackage{graphicx}
\usepackage{subfigure}
\usepackage{booktabs} 

\usepackage{hyperref}
\usepackage{balance} 

\usepackage[accepted]{icml2024}

\usepackage{amsmath}
\usepackage{amssymb}
\usepackage{mathtools}
\usepackage{amsthm}

\usepackage[capitalize,noabbrev]{cleveref}

\usepackage{multirow}
\usepackage{caption}
\usepackage{subcaption}

\usepackage{kotex}
\usepackage{booktabs} 
\usepackage{adjustbox}
\usepackage{makecell}

\usepackage{amssymb}
\usepackage{pifont}

\usepackage[capitalize,noabbrev]{cleveref}

\theoremstyle{plain}

\theoremstyle{definition}

\theoremstyle{remark}

\usepackage[textsize=tiny]{todonotes}

\icmltitlerunning{Self-Supervised Interpretable End-to-End Learning via Latent Functional Modularity}

\begin{document}

\twocolumn[
\icmltitle{
Self-Supervised Interpretable End-to-End Learning via \\Latent Functional Modularity
}




\begin{icmlauthorlist}
\icmlauthor{Hyunki Seong}{kaist}
\icmlauthor{David Hyunchul Shim}{kaist}
\end{icmlauthorlist}

\icmlaffiliation{kaist}{School of Electrical Engineering, Korea Advanced Institute of Science and Technology, Daejeon, South Korea}

\icmlcorrespondingauthor{Hyunki Seong}{hynkis@kaist.ac.kr}

\icmlkeywords{End-to-End Learning, Self-Supervised Learning, Interpretable Learning, Modular Network, Imitation Learning}

\vskip 0.3in
]

\printAffiliationsAndNotice 

\input{sections/0.abstract.tex}
\input{sections/1.introduction}
\input{sections/2.related_works}
\input{sections/3.methodologies}
\input{sections/4.experiments}
\input{sections/5.conclusion}


\section*{Acknowledgements}
We would like to thank the anonymous reviewers for their inspiring questions and feedbacks for our work. 

\section*{Impact Statement}
This paper aims to advance the fields of Machine Learning and Robotics. While our study has numerous potential societal impacts, we believe none require specific emphasis in this context.

\balance

\bibliography{main}
\bibliographystyle{icml2024}

\clearpage
\appendix
\input{sections/X_suppl}

\end{document}

%% file: sections/0.abstract.tex
\begin{abstract}
We introduce MoNet, a novel functionally modular network for self-supervised and interpretable end-to-end learning. By leveraging its functional modularity with a latent-guided contrastive loss function, MoNet efficiently learns task-specific decision-making processes in latent space without requiring task-level supervision. Moreover, our method incorporates an online, post-hoc explainability approach that enhances the interpretability of end-to-end inferences without compromising sensorimotor control performance. In real-world indoor environments, MoNet demonstrates effective visual autonomous navigation, outperforming baseline models by 7\% to 28\% in task specificity analysis. We further explore the interpretability of our network through post-hoc analysis of perceptual saliency maps and latent decision vectors. This provides valuable insights into the incorporation of explainable artificial intelligence into robotic learning, encompassing both perceptual and behavioral perspectives. Supplementary materials are available at \url{https://sites.google.com/view/monet-lgc}.
\end{abstract}

%% file: sections/1.introduction.tex
\section{Introduction}
\label{sec:introduction}
\begin{figure}[t]
\centering
\includegraphics[width=0.48\textwidth]{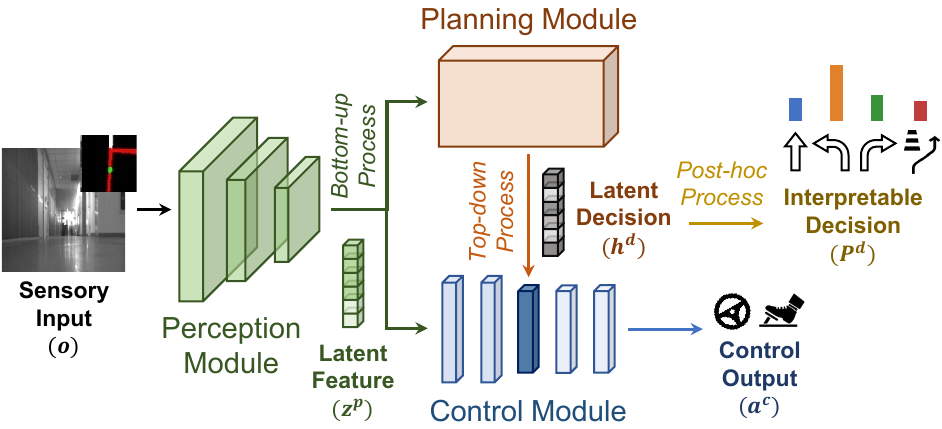}
\caption{
Our approach incorporates a functionally modular end-to-end network architecture, which includes a post-hoc method for an interpretable latent decision-making process.
}
\label{fig:intro}
\vspace{-5pt}
\end{figure}

One of the main objectives of end-to-end learning for autonomous navigation is to develop complex policies through human demonstrations. This is achieved by an end-to-end network that learns the hierarchical pipeline of perception, planning, and control in robotic systems via imitation learning (IL). Given that IL facilitates safe and efficient policy learning in an offline, supervised manner, end-to-end networks have been widely used in the design of learning-based applications~\cite{tampuu2020survey}.

However, although studies on IL have shown preliminary successes, designing an end-to-end sensorimotor network that can scale up to complex driving scenarios remains challenging.
Traditional end-to-end networks often exhibit a less clear decision-making process, which complicates learning entangled tasks from demonstrations. To address this, recent conditional learning methods~\cite{huang2020multi} employ multiple branching networks for each task, with outputs that switch based on task-level conditional inputs. However, this conditional input often corresponds to the outcome of an internal decision-making process in human demonstrations, which is typically implicit and difficult to identify. As a consequence, this approach necessitates extra task-level annotations (e.g., \textit{go-straight}, \textit{turn-left}), making it more demanding than simply collecting sensorimotor pairs.

Moreover, conventional networks, which directly compute control commands from sensory inputs, lack a transparent inference process. This obscurity makes it unclear what behavioral decision was intended for the resulting control output without direct execution. Several studies have sought to enhance interpretability by reconstructing various modalities~\cite{chen2021interpretable,zeng2019end}, or by visualizing attention maps~\cite{kim2017interpretable}. However, they mainly focus on perceptual insights, still leaving the high-level decisions behind sensorimotor outputs largely obscure. This lack of clarity leads to insufficient task specificity and interpretability during the sensorimotor process, ultimately diminishing the reliability and trustworthiness of end-to-end networks in practical applications.

In this paper, we present a modular network, MoNet, a functionally modularized end-to-end architecture~\cite{meunier2010modular} for autonomous navigation. MoNet is divided into perception, planning, and control modules, which are functionally separated but explicitly connected to form a single end-to-end structure (Fig. \ref{fig:intro}).
Our network includes an internal latent decision process to facilitate task-oriented guidance for behaviorally relevant sensory-motor processes. Simultaneously, it employs a self-attention mechanism to extract salient spatial features from sensory input.

Leveraging the modularity in MoNet, we design a novel self-supervised, latent-guided contrastive (LGC) loss function. Directed by latent features from the perception module with task-oriented contexts, this loss function encourages the planning module to make consistent decisions in similar driving contexts while differentiating responses in varied situations. The internal hierarchy, combined with our contrastive learning scheme, not only promotes functional specialization but also enables the emergence of a task-relevant decision-making mechanism through self-supervision.

Furthermore, we integrate a post-hoc technique from the field of explainable artificial intelligence (XAI) with our modular end-to-end network to transform task-relevant latent decisions into understandable representations. We implement a multi-class pattern classifier to predict the high-level task intent derived from these latent decisions. Subsequently, we calibrate the posterior probabilities of the classification results to achieve a more interpretable representation. These probabilities are then converted into an entropy value, which quantifies the uncertainty of the end-to-end model's inference from a task-level perspective.

In our evaluation, our method effectively demonstrates visual autonomous driving across multiple tasks, including corridor navigation, intersection navigation, and collision avoidance. We present empirical experiments conducted on a real-world robotic RC platform, showcasing the network’s capability to perform task-specific sensorimotor inference without requiring task-level labeling.
We further explore spatial saliency maps and latent decisions during end-to-end navigation in the real world. Specifically, by decoding latent decisions into explainable posterior probabilities, we gain the ability to visualize sequential high-level internal decisions alongside task uncertainty during continuous end-to-end sensorimotor control. These analyses highlight the significant interpretability and transparency of our end-to-end model, showcasing its effectiveness from both perceptual and behavioral perspectives in real-world continuous control applications.

Our main contributions can be summarized as follows:
\begin{itemize}
    \item
    We propose MoNet, a modular end-to-end network that incorporates a post-hoc explainability method, enabling interpretable sensorimotor control.
    \item
    We design a self-supervised, latent-guided contrastive learning scheme to enhance the task-relevant decision-making mechanism within the end-to-end architecture.
    \item
    We examine the perceptual and behavioral interpretability, as well as the sensorimotor performance of our network, showcasing the potential benefits of integrating the explainability method into robotic learning.
\end{itemize}

%% file: sections/2.related_works.tex
\section{Related Works}
\label{sec:related_works}
\paragraph{End-to-End Sensorimotor Learning:}
In autonomous driving, end-to-end methods employ single neural networks to directly map sensory inputs to control outputs. ALVINN, the initial model for steering angle inference, utilized a multilayer perceptron~\cite{pomerleau1988alvinn}. This approach has evolved to include convolutional neural networks (CNNs), mainly focused on lane-following tasks~\cite{bojarski2016end}. Recent advancements have incorporated conditional imitation learning to cover a broader range of driving tasks~\cite{gao2017intention, codevilla2018end,huang2020multi,zhang2023coaching}. These methods use multiple branched layers switched by conditional inputs for navigating environments, such as \textit{'go-straight'}, \textit{'turn-left'}, or \textit{'turn-right'}. While such methods reduce task-level ambiguity, they necessitate additional human-engineered labeling for the navigational inputs and are constrained to predefined tasks. Moreover, interpreting the perceptual and behavioral processes within end-to-end networks remains a challenge, which affects confidence in the network's reliability for real-world deployment.

\begin{figure*}[t]
\centering
\includegraphics[width=0.99\textwidth]{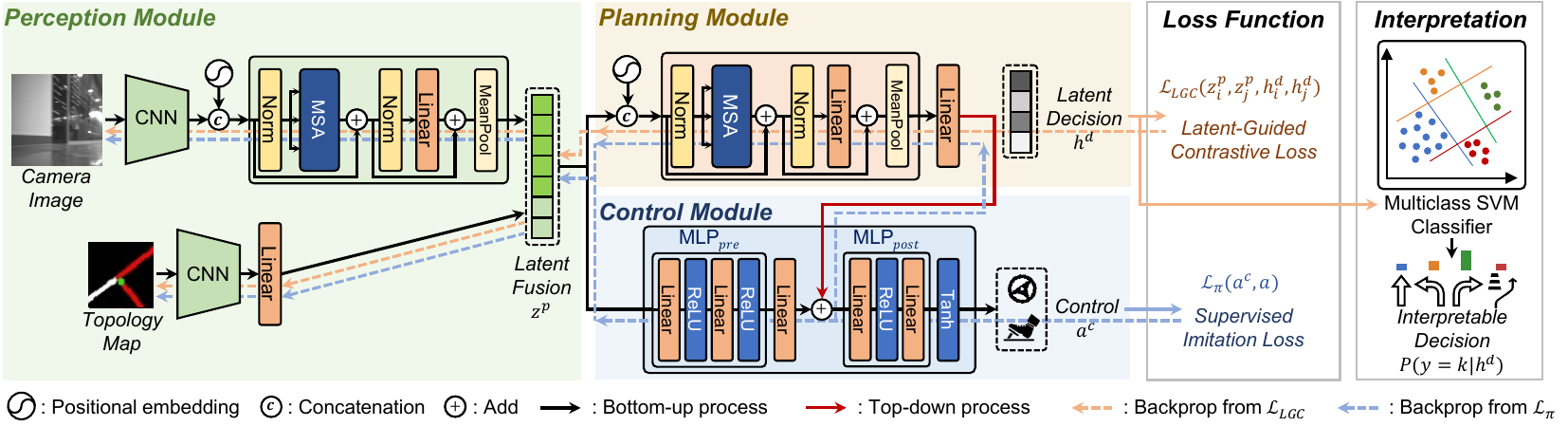}
\caption{
Overview of our method. While the entire end-to-end network is optimized by the supervised imitation loss $\mathcal{L}_{\pi}$, the planning module is updated by the latent-guided contrastive loss $\mathcal{L}_{LGC}$, which is directed by the latent vector $z^p$.
}
\label{fig:network}
\vspace{-10pt}
\end{figure*}

\paragraph{Interpretable Methods:}
Recent studies have concentrated on designing interpretable end-to-end networks to address existing limitations. In this context, researchers using segmentation methods~\cite{chen2021interpretable,teng2022hierarchical} have indirectly shown how a network can comprehend surrounding contexts by generating semantic masks from hidden features. Similarly, studies involving multi-head networks~\cite{zeng2019end} have evaluated the effectiveness of their planning methods by examining interpretable representations across various modalities, such as object detection or cost map generation. In contrast, attention mechanisms~\cite{vaswani2017attention,kim2017interpretable} in recent studies have explicitly facilitated a deeper understanding of the areas within given feature elements where the network predominantly focuses during feedforward processing. Specifically, in the realm of autonomous driving, methods leveraging attention aim to accentuate critical aspects in driving scenarios, such as lane following~\cite{shi2020self}, lane changing~\cite{chen2019attention}, or navigating intersections~\cite{seong2021learning}. However, the majority of research has primarily focused on the cognitive interpretations of how networks perceive contexts. Our work takes this a step further by investigating how to interpret the task-oriented intentions of the network in an explainable way. This approach enables both perceptual and behavioral interpretations online during end-to-end inference.

%% file: sections/3.methodologies.tex
\section{Modular End-to-End Network}
\label{sec:modular-network}
\subsection{Latent Functional Modularity}
Our main idea is to embed functional modularity with internal hierarchy into an end-to-end network, allowing functionalities of the robotic sub-modules in latent space.
As shown in Fig. \ref{fig:intro}, our modular end-to-end network, MoNet, has three distinct neural modules: \textit{Perception} ($\mathcal{P}$), \textit{Planning} ($\mathcal{Q}$), and \textit{Control} $(\mathcal{R})$, which are the major components of the robotics system~\cite{schwarting2018planning}.
Each module 1) encodes raw observations $o$ into a fused perception feature vector $z^p$, 2) infers a latent decision $h^d$, and 3) computes a sensorimotor command ${a}^c$, respectively. The modules are functionally separated yet structurally connected in latent space, enabling them to constitute an end-to-end policy network $\pi$, parameterized by $\theta$:
\begin{equation}
    \begin{aligned}
        \label{eq:modules_equations}
        &\textit{Perception: } &{z}^p  = \mathcal{P}({o}; \theta) \\
        &\textit{Planning: }   &{h}^d = \mathcal{Q}({z}^p; \theta) \\
        &\textit{Control: } &{a}^c = \mathcal{R}({z}^p, {h}^d; \theta) \\
    \end{aligned}
\end{equation}
To encourage functional specialization of the modules in the network, we utilize two distinct mechanisms: bottom-up and top-down neural processes ~\cite{baluch2011mechanisms, anderson2018bottom}. Specifically, the bottom-up mechanism is a stimulus-driven, exogenous process, while the top-down mechanism is a behavior-relevant endogenous process~\cite{katsuki2014bottom}. Considering their properties, the perception ($\mathcal{P}_{\theta}$) and planning ($\mathcal{Q}_{\theta}$) modules configure with self-attention mechanisms~\cite{vaswani2017attention} to extract salient spatial features from sensory input $o$ and to obtain contextual importance from the features $z^p$, respectively. In contrast, the control ($\mathcal{R}_{\theta}$) module is designed with a top-down mechanism to internally modulate the sensory-motor signals based on the context-oriented behavioral decision $h^d$ from the planning module.
This internal hierarchy enables the network to generate spatial attention maps and high-level latent decisions that are explicitly accessible during end-to-end inference. Employing a post-hoc approach allows these to be transformed into interpretable salient maps and behavioral decisions, respectively.

\subsection{Network Details}
\paragraph{Perception module:}
Our network receives a high-dimensional observation ${o} = [I, M]$ that includes a front camera image $I \in \mathbb{R}^{224\times224\times1}$ and a topology map $M \in \mathbb{R}^{64\times64\times3}$. This observation includes visual sensory data with navigational information, providing driving contexts in the ego-centric area for navigating complex environments, such as corridors with intersections.
To effectively process high-dimensional camera images, we employ a hybrid architecture that combines the Vision Transformer encoder with CNN blocks~\cite{dosovitskiy2020vit}.
In the perception module, the image $I$ and the topology map $M$ are first encoded into a hidden image feature $z_I \in \mathbb{R}^{6\times6\times64}$ and a hidden route feature $z_M \in \mathbb{R}^{1\times1\times64}$, respectively, using ResNet-inspired CNN blocks~\cite{he2016deep}.
The image feature is first reshaped into a flattened embedding $z^{\text{flat}}_{I} \in \mathbb{R}^{(6 \times 6) \times 64}$, serving as a tokenized embedding for $N\!=6\!\times\!6$ image patches. This reshaped embedding is then concatenated with a positional embedding and fed into a Transformer encoder network (see Appendix \ref{appendix:perception}). Subsequently, the Transformer model processes this input to produce an attention matrix $A(\cdot,\cdot)$, which integrates with the global context of the image feature $z_I$ via the self-attention mechanism.
The attention matrix is computed as follows:
\begin{flalign}
    \label{eq:self_attention}
    A(Q,K) = softmax(\frac{QK^T}{\sqrt{d_k}}) \\
    \label{eq:QKV}
    Q, K, V = Z W_{Q}, Z W_{K}, Z W_{V}
\end{flalign}
where $Q, K, V \in \mathbb{R}^{N \times D_k}$ refer to queries, keys, and values consisting of $N$ data nodes with $D_k$ dimension size by following common terminology~\cite{dosovitskiy2020vit}. $W_{Q}, W_{K}, W_{V} \in \mathbb{R}^{D_k \times D}$ are weight matrices for an arbitrary input feature $Z \in \mathbb{R}^{N \times D}$ with $D$ dimension size to compute $Q, K, V$.
The attention matrix $A \in \mathbb{R}^{N \times N}$ discerns the spatial significance of feature elements in the input image, offering valuable information to interpret the module's bottom-up neural processing from a perceptual standpoint.
Finally, by applying mean pooling, we derive the attention-integrated feature $z^{\text{att}}_I \in \mathbb{R}^{65}$ with reduced dimensionality. This feature is then concatenated with the flattened route feature $z^{\text{flat}}_{M} \in \mathbb{R}^{64}$, yielding a latent feature fusion vector $z^p = [z^{\text{att}}_I, z^{\text{flat}}_{M}]$. The fused feature is then fed into the planning and control modules without nonlinearity.

\paragraph{Planning module:}
This module extracts contextual features from the fused vector $z^p$ and produces a latent decision $h^d$ to modulate the neural signals of the control module in a top-down manner.
We construct the planning module using another Transformer network, mirroring the encoder of the perception module. Here, the fusion vector is expanded and tokenized into an input embedding $z^{emb} \in \mathbb{R}^{(65+64)\!\times\!64}$, corresponding to the embedding $z^{\text{flat}}_{I}$ in perception. This input embedding is then concatenated with a positional embedding, following the same process as that in the perception module. To derive the latent decision in continuous space, we apply a linear layer to the output of the Transformer encoder without using a nonlinear activation function.

\paragraph{Control module:}
The control module computes a low-level control command incorporating the high-level decision through bottom-up and top-down processes (Fig. \ref{fig:network}).
The module initially extracts a pre-sensory-motor feature $x_{pre} \in \mathbb{R}^{N_d}$ from a given perceptual feature $z^p$ using the $\text{MLP}_{pre}$ block (Eq. \ref{eq:control-pre-mlp}). The motor feature is then passed through a linear fully-connected layer (FC) and modulated in a top-down fashion via elementwise addition with the task-oriented latent decision $h^d \in \mathbb{R}^{N_d}$ (Eq. \ref{eq:control-top-down}). The module finally converts the modulated feature into the control command $a^c \in \mathbb{R}^{2}$ through the $\text{MLP}_{post}$ block followed by a tanh activation function (Eq. \ref{eq:control-post-mlp}).
\begin{flalign}
    \label{eq:control-pre-mlp}
    x_{pre} &= \text{MLP}_{pre}(z^p) \\
    \label{eq:control-top-down}
    x_{mod} &= \text{FC}(x_{pre}) + h^d \\
    \label{eq:control-post-mlp}
    a^c &= \text{tanh}(\text{MLP}_{post}(x_{mod}))
\end{flalign}
Here, the command $a^c = [\delta^c, \tau^c]$ contains a normalized steering angle $\delta^c$ and throttle value $\tau^c$.
This self-modulated hierarchy facilitates the independent computation of sensorimotor and contextual data from perceptual inputs, resulting in a control signal guided by latent decision-making. As a result, MoNet is capable of learning task-specific sensorimotor policies even from task-agnostic demonstrations.

\begin{figure}[t!]
\centering
\includegraphics[width=0.45\textwidth]{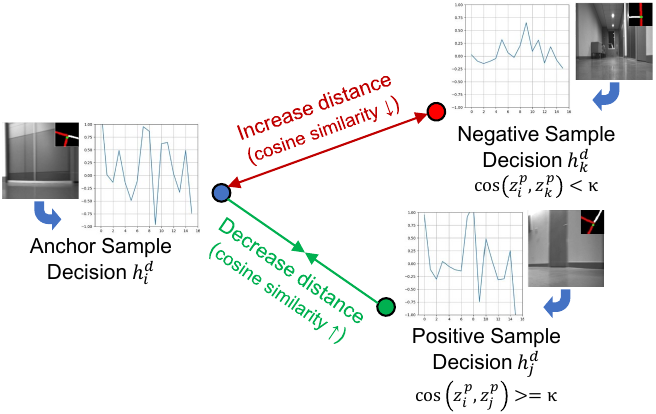}
\caption{
Our self-supervised contrastive learning scheme assesses the similarity of the perceptual features to decide on positive and negative latent decision samples.
}
\label{fig:LGC}
\vspace{-15pt}
\end{figure}
\subsection{Training Details}
To train the network, we first introduce the supervised loss function $\mathcal{L}_{\pi}$, defined as the absolute deviation (L1) between the network's prediction and the demonstration data:
\begin{flalign}
    \label{eq:loss-imit}
    \mathcal{L}_{\pi}({a}^c, {a}) = \lvert \delta^c - \delta \rvert + \lambda_{\tau} \lvert \tau^c - \tau \rvert
\end{flalign}
where the loss term for throttle control $\tau^c$ is weighted by the parameter $\lambda_{\tau} \in [0, 1]$. This weighting aims to emphasize supervision on steering control in visual autonomous navigation~\cite{codevilla2018end}. Given that we collect noisy demonstrations from a real robot platform, we choose the L1 loss to reduce the penalty for large errors and be more robust to outliers compared to the L2 loss.

\begin{figure*}[t!]
\centering
\includegraphics[width=0.95\textwidth]{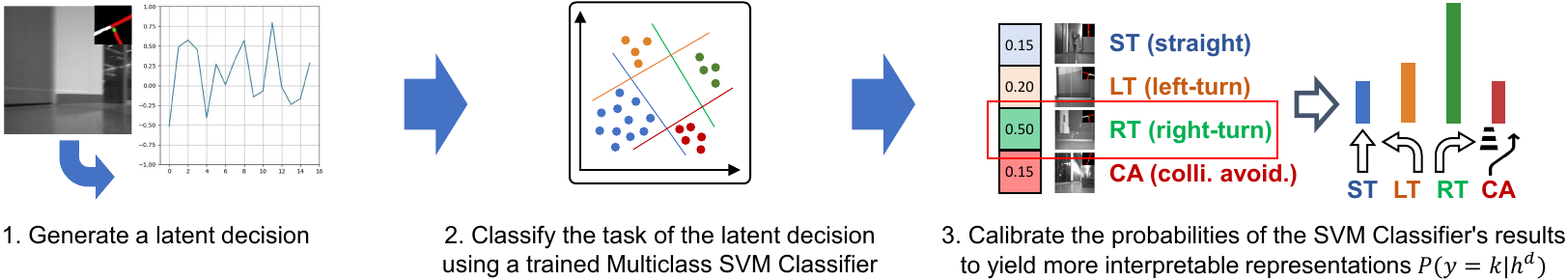}
\caption{Overview of our post-hoc behavior interpretation process.
}
\label{fig:posthoc}
\vspace{-15pt}
\end{figure*}
Furthermore, to enhance the distinctiveness of top-down latent decisions, we design a latent-guided contrastive (LGC) loss function using a self-supervised approach, leveraging the modular characteristics of our end-to-end network (Fig.~\ref{fig:LGC}). Generally, the output of planning is influenced by the context of the driving scene. This implies that similar situations lead to analogous decisions, while different scenarios result in distinct plans. Building on the observation that planning outputs are context-dependent, we define the latent-guided contrastive loss, denoted as $\mathcal{L}_{\text{LGC}}({z}^p_i, {z}^p_j, {h}^d_i, {h}^d_j)$. Here, the latent decision $h^d$ is guided by the feature fusion $z^p$ of the perception module as follows:
\begin{flalign}
    \label{eq:loss-cont}
    \mathcal{L}_{\text{LGC}} = 
        \begin{cases}
        \hfil 1 - \text{cos}({h}^d_i, {h}^d_j) & \!\!\!\!\text{if} \ \text{cos}({z}^p_i, {z}^p_j) >= \kappa  \\
        \hfil \text{max}(0, \text{cos}({h}^d_i, {h}^d_j)) & \!\!\!\!\text{if} \ \text{cos}({z}^p_i, {z}^p_j) < \kappa
        \end{cases}
\end{flalign}
where $\text{cos}(\alpha,\beta) = \frac{\alpha \cdot \beta}{|\alpha||\beta|}$ is cosine similarity that is widely used for similarity and clustering analysis in data science~\cite{larose2014discovering}.
It calculates the distance between two vectors based on their relative orientations, rather than their absolute distance, within the bounded range [-1, +1]. The subscript $j$ represents the index of a sample within a mini-batch, different from the current sample index $i$.
By minimizing Eq. \ref{eq:loss-cont}, we aim to reduce the intra-cluster distance for latent decisions in demonstrations where perceptual feature similarity exceeds $\kappa$. Conversely, we strive to increase the inter-cluster distance when the similarity is below $\kappa$. This approach incentivizes the planning module to generate more consistent latent decisions for scenarios with comparable perceptual contexts, while ensuring diverse decisions for scenarios with differing contexts. Moreover, by leveraging perceptual features, our method eliminates the need to define positive or negative samples, thereby enabling contrastive learning through a self-supervised approach.

The overall per-sample loss function is given by the weighted summation:
\begin{flalign}
    \label{eq:loss-per-sample}
    \mathcal{L} = \mathcal{L}_{\pi}({a^c_i}, {a_i}) + \lambda_{\text{LGC}} \mathcal{L}_{\text{LGC}}({z}^p_i, {z}^p_j, {h}^d_i, {h}^d_j)
\end{flalign}
where $\lambda_{\text{LGC}}$ is a weight parameter.
During the training phase, the supervised loss function $\mathcal{L}_{\pi}$ propagates the gradient flow across all modules ($\mathcal{P,Q,R}$), while the latent-guided loss function $\mathcal{L}_{\text{LGC}}$ targets only the planning and perception modules ($\mathcal{P,Q}$), promoting functional distinction between the planning and control modules ($\mathcal{Q,R}$).

\subsection{Interpretation Details}
\paragraph{Perceptual Interpretation:}
We use the attention matrix of the perception module to create a saliency map $S$, which highlights the spatial regions in the current driving scene that the network focuses on from a perceptual perspective.
The module generates the attention matrix $A \in \mathbb{R}^{N \times N}$ corresponding to the flattened vector of the encoded feature map $z_I \in \mathbb{R}^{h \times w \times c}$, where $N = h \times w$ is a resulting size of attention, $(h, w)$ is a reduced resolution of the image $I \in \mathbb{R}^{224 \times 224 \times 1}$, and $c$ is the feature dimension of $z_I$. Thus, we initially aggregate weights along the first dimension of $A$ to obtain the averaged attention weights $\bar{A} \in \mathbb{R}^{1 \times N}$:
\begin{flalign}
    \label{eq:saliency_map}
    \bar{A}_{j} = \frac{1}{N}\sum\nolimits_{i=1}^{N} A_{ij} \quad \textrm{for} \quad j = 1, ..., N
\end{flalign}
where $\bar{A}_{j}$ represents the central tendency of the weights in each column. Subsequently, we reshape the averaged weights into a two-dimensional matrix $\bar{S} \in \mathbb{R}^{h \times w}$. This is then upscaled to form the saliency map $S \in \mathbb{R}^{224 \times 224}$.

\paragraph{Behavioral Interpretation:}
Considering that the latent decision contains task-oriented features, we decode the decision vector $h^d$ into an understandable, task-wise probability score vector to facilitate the behavioral interpretation of our network.
We employ a multiclass linear Support Vector Machine (SVM) classifier~\cite{suthaharan2016machine} that is computationally efficient and less prone to overfitting. Utilizing sample decisions and their corresponding task labels $(h^d_i, y_i)$, linear SVM is designed to learn binary classification through the following optimization~\cite{tang2013deep}:
\begin{gather}
    \label{eq:svm}
    \min_{w, b} \frac{1}{2} w^T w\!+ C \sum_{i=1}^{M} max(0, 1 \!- y_i(w^T h^d_i + b))^2
\end{gather}
where $w$ is the weight vector, $b$ is the bias, and $C$ is the regularization parameter.
The SVM is extended to multiclass classification using the one-vs-rest scheme. This adaptation enables SVMs to maximize the margin between input data belonging to different classes.
After training the multiclass SVMs, we transform their output into a posterior probability score vector $P(y_i\!=k|h^d_i)$ for each class $k$, using a calibration method~\cite{niculescu2005predicting}:
\begin{flalign}
    \label{eq:decision_prob}
    P(y_i=k|h^d_i) = \frac{1}{1 + exp(E_k f_k(h^d_i) + F_k)}
\end{flalign}
where $f_k=w_k^T h^d_i + b_k$ is the SVM's output for class $k$, and $E_k$, $F_k$ are parameters fitted using maximum likelihood estimation from sample data set $[f_k(h^d_i), y_i]$.

We carry out behavior interpretation in a post-hoc manner. Initially, we generate sample latent decisions for each task that human engineers aim to interpret, using the trained modular network. Subsequently, we train the multiclass SVMs, along with parameters $E_k$ and $F_k$ for the calibration method. Finally, during sensorimotor inferencing, we transform MoNet's latent decisions into score vectors using Eq.~\ref{eq:decision_prob} (Fig.~\ref{fig:posthoc}). This approach allows us to interpret the end-to-end model without sacrificing sensorimotor performance.

Our approach is comparable to concept-based interpretation methods in explainable artificial intelligence~\cite{ghorbani2019towards}. These studies focus on understanding how high-level concepts are represented and utilized by models in decision-making. In our case, the concept vector corresponds to the latent decision that encapsulates the \textit{driving situation}. Consequently, to interpret the decision intent during the sensorimotor process, we quantify the alignment of a given decision vector with the specific tasks' concept (driving situation). This is performed by decoding the latent decision into the understandable posterior probabilities.

%% file: sections/4.experiments.tex
\section{Experiments}
\label{sec:experiments}
\begin{figure}[t]
  \centering
  \includegraphics[width=0.48\textwidth]{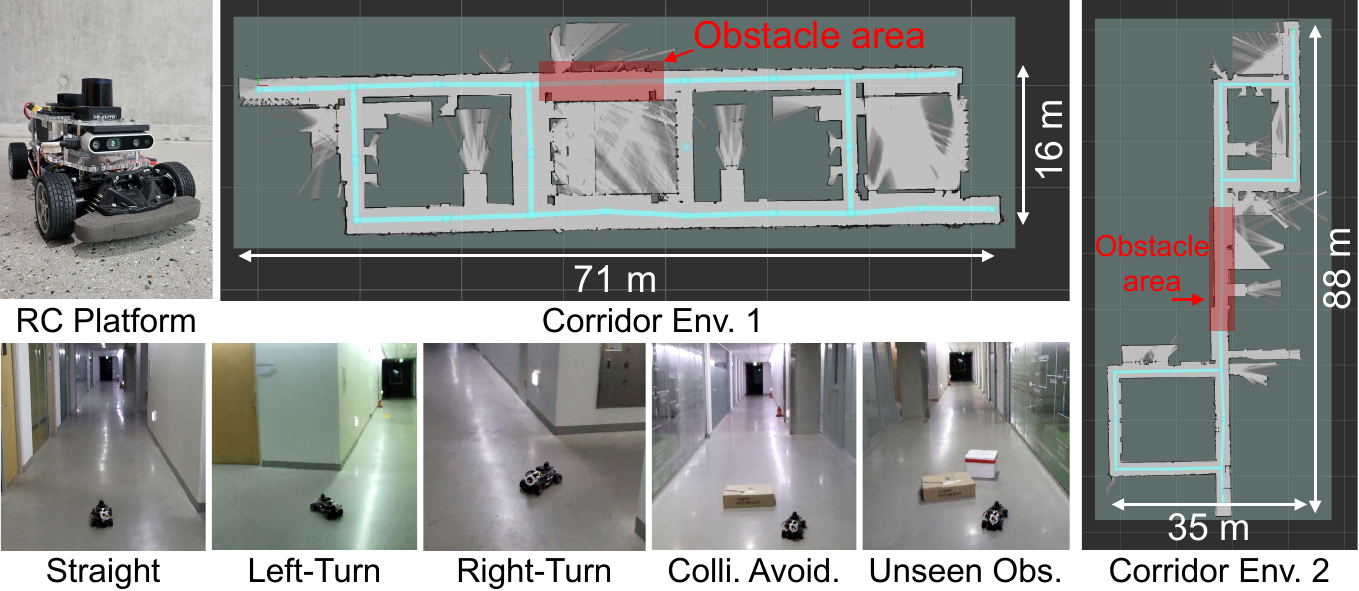}
  \caption{
  Hardware and experimental setup.
  }
  \label{fig:exp_setup}
  \vspace{-20pt}
\end{figure}

\begin{figure*}[t!]
\centering
\includegraphics[width=0.99\textwidth]{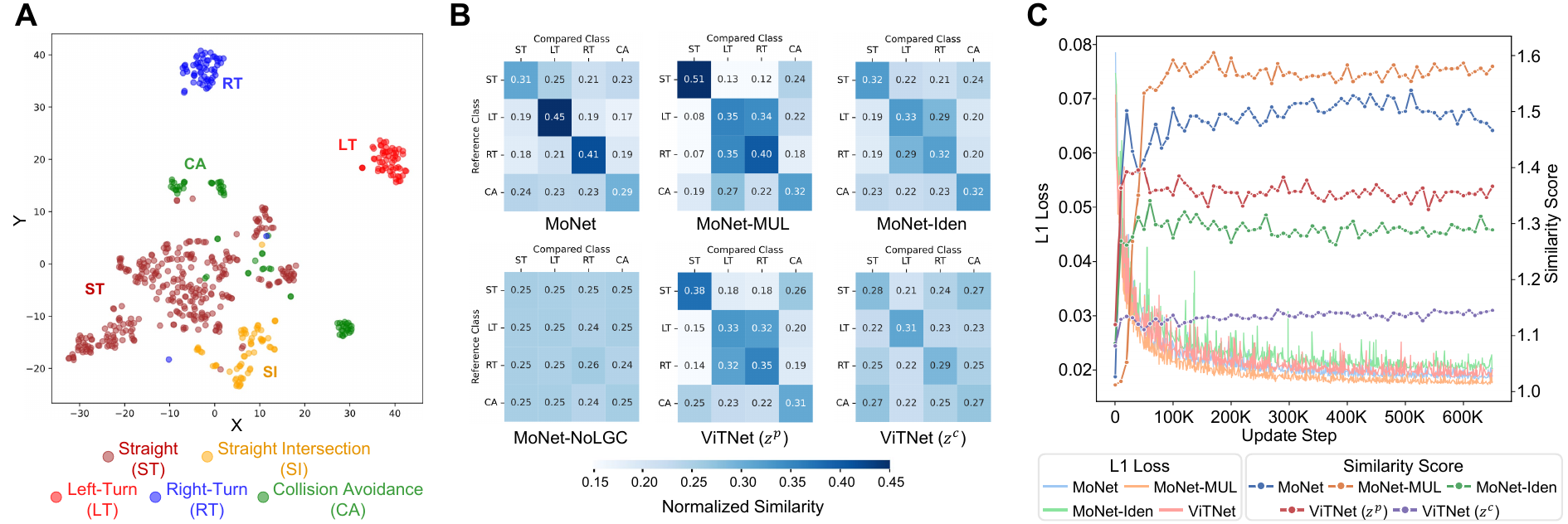}
\caption{Results of the planning-level quantitative evaluations. \textbf{(A): }The t-SNE map illustrates clusters of latent decisions among different tasks. \textbf{(B): }The similarity matrix shows the quantitative similarity of the decision vectors between different classes. \textbf{(C): }The learning curve indicates performance improvement in L1 loss and similarity score across different models.
}
\label{fig:result_quantitative}
\vspace{-10pt}
\end{figure*}

\subsection{Experimental Setup}
Fig. \ref{fig:exp_setup} shows the overall setup for hardware, environment, and scenarios in this work.
We apply MoNet on a wheeled, car-like platform, modeled after the F1TENTH vehicle~\cite{o2020f1tenth}. Our platform consists of a 1/10 scale racing car chassis (TT-02) equipped with an embedded computer (Jetson Xavier NX) and a controller (Arduino).
The Xavier NX receives front camera images and range measurements from sensors mounted on the platform. The range measurements are utilized to estimate the current pose of the ego vehicle and compute ego-centric coarse topology map~\cite{amini2019variational}. Detailed hardware setup and the coarse map processing are provided in Appendix~\ref{appendix:hardware}.

Our platform performs visual autonomous navigation with multiple driving tasks such as \textit{straight} (ST), \textit{straight-intersection} (SI), \textit{left-turn} (LT), \textit{right-turn} (RT), and \textit{collision avoidance} (CA).
Data collection is carried out under controlled conditions in two indoor environments: Corridor Environment 1 (Env. 1, $71m \times 16m$) and Environment 2 (Env. 2, $88m \times 35m$). Box-shaped obstacles are randomly positioned within specific areas in these environments. The training dataset comprises data from scenarios that feature either a single obstacle or no obstacles. However, scenarios involving multiple obstacles are introduced as new, unseen challenges during the evaluation phases. Our method is evaluated in Env. 1, characterized by more frequent intersection situations during autonomous navigation. For further details on data collection and processing, we refer to Appendix~\ref{appendix:data}.

\subsection{Quantitative Evaluation}
\paragraph{Baseline Models}
In addition to our method, we have implemented ViTNet, a baseline model designed as a Vision Transformer-based end-to-end architecture comprising only perception ($\mathcal{P}$) and control ($\mathcal{R}$) modules. This design allows us to investigate the necessity of the planning module ($\mathcal{Q}$).
For comparison with the latent decision, we select the perceptual ($z^p$) and the control-level hidden features ($z^c$) of ViTNet. Here, $z^c$ is the output of Eq. \ref{eq:control-top-down}, computed without involving the internal decision process.
Additionally, we introduce MoNet-based methods, MoNet-MUL, MoNet-Iden, and MoNet-NoLGC, for the ablation study. To analyze latent decision computation, MoNet-MUL is configured to perform element-wise multiplication instead of an additive process in Eq.\ref{eq:control-top-down}. MoNet-Iden, whose planning module acts as an identity function, is developed to assess the impact of neural processing within the planning module. MoNet-NoLGC is trained without the LGC loss function to investigate the impact of our self-supervised contrastive learning approach on the task specificity of the network.

\paragraph{Planning Performance}
To assess the planning-level performance of the end-to-end network, we quantify the task specificity during sensorimotor inference using a t-SNE map~\cite{van2008visualizing} and a Representational Similarity Matrix (RSM)~\cite{popal2019guide}. These analyses provide the user with a clear understanding of the network's performance in discriminating between different tasks.
In consideration of the data distribution, we sampled 318, 64, 59, 59, and 67 pieces of data, respectively, for the five tasks (ST, SI, LT, RT, CA) for these assessments.

The t-SNE visualization (Fig.~\ref{fig:result_quantitative} (A)) demonstrates that our planning module generates distinct and well-structured decisions in the latent space for various tasks. It effectively differentiates between the LT and RT tasks from the ST scenario and recognizes the directional variations in navigating intersections.
The decisions for CA are positioned between the ST, LT, and RT clusters, indicating a need for moderate planning that involves both intersection-turning and corridor-following behaviors in collision avoidance scenarios.
Furthermore, the data for SI exhibits a high similarity to that of ST, reflecting similar driving contexts, despite their differing inputs from topological maps.
This result highlights that our network is adept at capturing the common driving context found in straight driving, whether it occurs in corridors or intersections.
Given these findings, and to ensure a clearer distinction of task classes, we have decided to classify ST and SI as the same task, designated as ST, in the experiments discussed later in this manuscript.

For a more quantitative analysis of latent planning, we further examine the RSM of the learned decisions across different baseline models (Fig.~\ref{fig:result_quantitative} (B)). We measure the cosine similarity between the latent decisions of each task using the average linkage method. The similarities among the four classes are then normalized to a range of [0, 1] through a row-wise softmax operation.
The results demonstrate that MoNet effectively differentiates between various tasks while clustering similar situations. The matrix shows strong diagonal values, indicating that the latent decisions effectively distinguish various driving tasks based on contextual features derived from sensory data, without task-level inputs. 
While the three models---MoNet-MUL, MoNet-Iden, and ViTNet($z^p$)---show high similarity values sufficient to separate multiple tasks, they struggle to distinguish between the directional characteristics of LT and RT. ViTNet($z^c$), which relies solely on control-level features, fails to represent discriminative task specificity among the four tasks.
Interestingly, in the absence of our LGC loss, the modular network MoNet-NoLGC also fails to learn task-specific features, reminiscent of the '\textit{collapse}' issue mentioned in~\cite{mittal2022modular}. This underscores the effect of our LGC loss in preventing collapse, significantly increasing the specialization of latent planning within the end-to-end network, even in a self-supervised manner.

\paragraph{Learning Curves}
We evaluate the learning curves of the baseline models based on their control and planning performance (Fig.~\ref{fig:result_quantitative} (C)). For control performance, we calculate the L1 loss using the validation dataset. To assess planning-level performance, we compute a similarity score, which is the sum of the diagonal values in the RSM results.
We skip the model MoNet-NoLGC because it does not show a meaningful similarity score compared to other baseline models (near 1.0).
Our approach achieves notable improvement in latent planning over other models, without compromising sensorimotor learning capabilities.
The L1 loss curves show minimal changes with the addition of an extra planning module or a contrastive learning scheme. This indicates that our method substantially enhances the task-specificity of end-to-end inferences without affecting policy learning.
Meanwhile, in the similarity score curves, our method outperforms other approaches, demonstrating 7\%-28\% higher final performance than ViTNet-based methods.
The latent decision ($h^d$) of MoNet, utilizing the self-supervised contrastive scheme, achieves a terminal score of 1.47, outperforming the perception-level (1.37) and control-level (1.15) hidden features from ViTNet.
Even when utilizing the identity function, the latent decision-making of MoNet-Iden shows better improvement in task specificity (1.29) compared to the control-level features.
This reveals that our latent decision-based approach embeds contextual characteristics more effectively compared to perceptual or low-level control features.
Although MoNet-MUL achieves the highest terminal score (1.58), based on the results of the RSM analysis, we have chosen MoNet with the additive process as our primary approach. This approach is selected for its ability to clearly address the four multiple driving tasks.

\begin{figure*}[t!]
\centering
\includegraphics[width=0.98\textwidth]{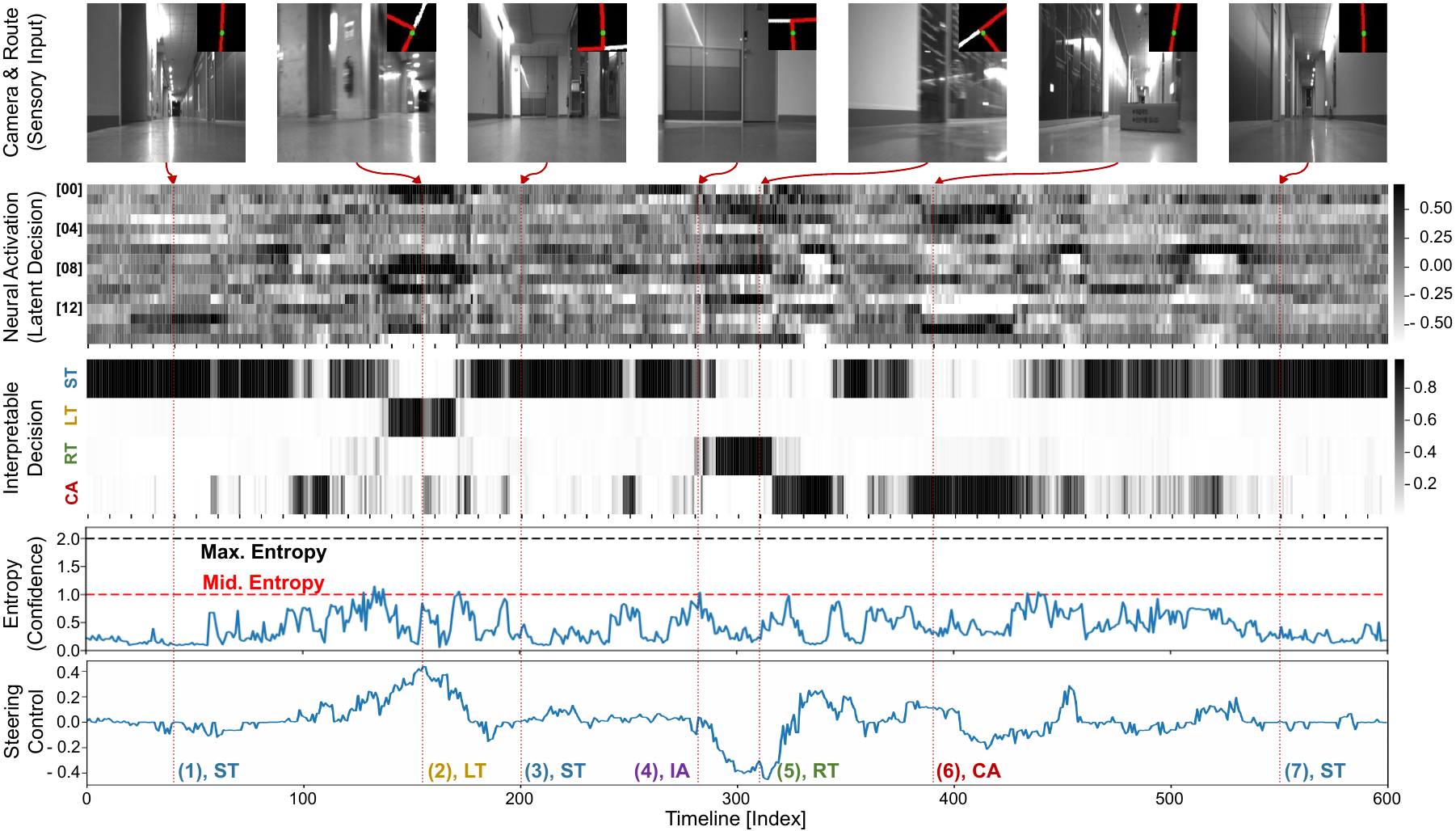}
\caption{Quantitative results showing given sensory inputs (front camera images with topological maps), latent decisions, decoded interpretable decisions, entropy, and control output during an autonomous navigation episode.
}
\label{fig:rollout}
\vspace{-10pt}
\end{figure*}

\paragraph{Sensorimotor Performance}
We evaluate the sensorimotor performance of MoNet by measuring the success rate of each task within the evaluation environment, comparing it with ViTNet, which features a perception-control-based end-to-end architecture.
Under the same hardware and environmental conditions, each model performed 16 episodes in the real-world environment, totaling 143 driving tasks. Table~\ref{tab:performance} summarizes the performance results. These results demonstrate that our model exhibits stronger generalization ability across multiple sensorimotor tasks compared to the baseline model. Both models show safe navigation performance in straight driving scenarios. However, ViTNet often struggles to overcome unseen obstacle scenarios and particularly fails in turning right at intersections, where it records its lowest success rate of 63\%. Although there was a situation where our model had a mild touch with a wall while avoiding cluttered obstacles, MoNet succeeded in all trials of navigating intersections and generally performed well in obstacle avoidance scenarios.
\begin{table}[t!]
  \centering
  \begin{adjustbox}{width=0.35\textwidth}
  \renewcommand{\arraystretch}{0.1}
  \begin{tabular}{@{}cccccc@{}}
    \toprule
    &\multicolumn{5}{c}{{Success Rate (Count/Total)}} \\
    \cmidrule(lr){2-6}
    Method & ST & SI & LT & RT & CA \\
    \midrule
    MoNet  & \makecell{1.00 \\ (76/76)} & \makecell{1.00 \\ (32/32)} & \makecell{1.00 \\ (8/8)} & \makecell{\textbf{1.00} \\ \textbf{(8/8)}} & \makecell{\textbf{0.95} \\ \textbf{(18/19)}} \\
    ViTNet & \makecell{1.00 \\ (76/76)} & \makecell{1.00 \\ (32/32)} & \makecell{1.00 \\ (8/8)} & \makecell{0.63 \\ (5/8)} & \makecell{0.89 \\ (17/19)} \\
    \bottomrule
  \end{tabular}
  \end{adjustbox}
  \caption{Success rate results for each driving task.}
  \label{tab:performance}
  \vspace{-10pt}
\end{table}

\subsection{Analysis of Interpretability}
We investigate the interpretability and transparency of our model while performing end-to-end sensorimotor processing by decoding the top-down latent decisions.
Fig.~\ref{fig:rollout} illustrates the quantitative results, including latent decisions, decoded interpretable decisions, and control output, during an autonomous navigation episode encompassing multiple tasks. Since the decoded decisions are represented as probabilistic score vectors, we further compute the entropy of these decisions. This entropy represents the confidence level of the internal decision-making during end-to-end processing.
The results show that our method can provide interpretable sensorimotor processes through decoded decisions that validly reflect the driving situation based on sensory inputs.
In the early phase, our robot followed a straight corridor using minimal steering control, demonstrating strong probability scores for the task decision ST. However, when navigating intersections, our model produced latent decisions that were decoded as high scores for corresponding turns (LT, RT), necessitating large steering commands. In the case of approaching a wall or obstacle, our network generates different patterns of neural activations (CA), resulting in unique decision responses compared to straight driving and turns (LT, RT). These results highlight that our approach enables a novel investigation of latent decision transitions during end-to-end inferences, thereby enhancing the transparency of online sensorimotor processes.

Moreover, by analyzing the entropy of the probability score vector, we can assess the confidence level of internal decision-making during end-to-end control. Whenever the robot needed to alter its current driving decisions, such as when approaching intersections or obstacles, the entropy of the decision increased to more than 1.0, indicating mid-level uncertainty values. Since latent decision-making is the causal process leading to robot control, our method can provide the internal confidence of the end-to-end inference prior to executing the robot's actions. This shows the significant interpretability of our model from a behavioral perspective in practical, real-world applications.

We further delve into perceptual and behavioral interpretation across various tasks by visualizing spatial saliency maps and interpretable decisions. We include these supplementary results in Appendix~\ref{appendix:decoding} for brevity.

%% file: sections/5.conclusion.tex
\section{Conclusion and Limitations}
\label{sec:conclusion}
We introduced MoNet, a modular network for self-supervised and interpretable end-to-end learning. Our method leverages functional modularity to enable a novel latent-guided contrastive learning scheme. This scheme allows the network to learn task-specific sensorimotor control without the need for task-level supervision. Furthermore, our network incorporates a self-attention mechanism and an internal decision process, both of which can be decoded into a spatial saliency map and an explainable decision. In real-world autonomous navigation, our model demonstrates effective sensorimotor performance with interpretability among multiple driving tasks.

Our approach to interpretable end-to-end learning with functional modularity offers several advantages for the use of end-to-end network architectures. Firstly, it enables more reliable and less uncertain end-to-end processes in robotics. Our method allows human engineers to comprehend the network's intent and the rationale behind control outputs from perspectives beyond control-level observation, including perception and planning. Such enhancement is particularly valuable in real-world deployments where safety is critical.
Secondly, our approach facilitates the integration of learning-based, black-box modules with nonlearning-based, white-box ones into a hybrid architecture. By leveraging decoded interpretable decisions from our modular network, it becomes feasible to conditionally apply either network-based policies or conventional controllers during deployment.
We hope our work contributes to integrating explainable artificial intelligence with end-to-end learning schemes, thereby enhancing the interpretability and transparency of learning-based robotic applications.

While MoNet shows promising results in real-world indoor environments, our method needs further extension to navigate more complex and dynamic environments, such as outdoor scenarios. Since these scenarios contain dynamic and varying features (e.g., moving objects, brightness), temporal features are crucial. An avenue for future work is to incorporate temporal network layers, such as LSTM, into our Vision Transformer-based perception module to learn temporally consistent features in dynamic driving scenes. We believe such a spatio-temporal module will enable our method to capture distinct task-level features with temporal consistency from a perceptual feature perspective. This will be one of the primary focuses of our future work.

%% file: sections/X_suppl.tex
\balance




\section{Supplementary Materials}
A demo video, codes, and dataset for quantitative results and interpretation are available at \url{https://sites.google.com/view/monet-lgc}.

\section{Method Details}
\subsection{Perception Module in Details}
\label{appendix:perception}
The perception module utilizes the Transformer encoder to generate a saliency map, which is then integrated with the global context of the input image $I$ through the self-attention mechanism.
Following the description in~\cite{dosovitskiy2020vit}, the encoder network comprises: 1) a multi-head self-attention block (MSA), and 2) an MLP block, both equipped with layer normalizations (LN) and residual connections. After feature extraction by the CNN block (Eq. \ref{eq:percept-cnn}), the input embedding undergoes preprocessing (Eq. \ref{eq:percept-embed}-\ref{eq:percept-flat}) before being fed into the Transformer encoder process (Eq. \ref{eq:percept-transformer-msa}-\ref{eq:percept-transformer-meanpool}).
\begin{flalign}
    \label{eq:percept-cnn}
    z_i &= \text{CNN}_i (o_i), \quad i=\{I, M\} \\
    \label{eq:percept-embed}
    z^{\text{flat}}_i &= \text{reshape}_i (z_i), \quad z^{\text{flat}}_I \in \mathbb{R}^{N_p \times D_p}, z^{\text{flat}}_M \in \mathbb{R}^{D_p} \\
    \label{eq:percept-flat}
    z^{I}_0 &= [z_I; z^{pos}_I] \in \mathbb{R}^{N_p \times (D_p+1)}, \quad z^{pos}_I \in \mathbb{R}^{N_p \times 1} \\
    \label{eq:percept-transformer-msa}
    {z}'_l &= \text{MSA}(\text{LN}({z}_{l-1})) + {z}_{l-1}, \quad l = 1, ..., L \\
    \label{eq:percept-transformer-mlp}
    {z}_l &= \text{MLP}(\text{LN}({z}'_l)) + {z}'_l, \quad l = 1, ..., L \\
    \label{eq:percept-transformer-meanpool}
    {z}^p &= [z^{\text{att}}_I; z^{\text{flat}}_M], \quad \text{where} \ z^{\text{att}}_I = \text{MeanPool}(z_L)
\end{flalign}

where $N_p = 6\times6$ and $D_p = 64$. The MLP block includes ReLU for nonlinearity. Considering the limited computing resources available for on-board implementation, we construct a single-stack Transformer encoder ($L=1$) for each module.
\begin{table}[h]
    \centering
    \begin{tabular}{cc}
    \toprule
         Batch size& 512\\
         Total training iterations& 650k\\
         Optimizer& Adam\\
         Similarity factor $\kappa$ & 0.5\\
         Weight for the LGC loss term $\lambda_{LGC}$ & 5e-4\\
         Learning rate& 3e-4\\
         Learning rate scheduler& LambdaLR\\
         Scheduler factor & 3e-4\\
    \bottomrule
    \end{tabular}
    \caption{Hyperparameter configuration}
    \label{tab:hyperparam}
\end{table}


\section{Experimental Details}
\subsection{Hyperparameter Setting}
\label{appendix:hyperparam}
Table~\ref{appendix:data} shows the hyperparameter setting for our experiments. The batch size is 512, and the total training iteration is 650k. We use Adam optimizer~\cite{kingma2014adam} with an initial learning rate of 3e-4. For self-supervised learning, we set the similarity factor $\kappa$ to 0.5 to ensure that the LGC loss function conservatively identifies positive samples in the early phase of training. 


\subsection{Hardware System and Coarse Topology Map}
\label{appendix:hardware}
Our platform consists of a 1/10 scale racing car chassis equipped with an embedded computer, Jetson Xavier NX, and a microcontroller, Arduino Nano. The Xavier NX receives front camera images with the Realsense D435i camera sensor and acquires range measurements using a 2D LiDAR sensor (Hokuyo UST-20LX). These measurements are utilized to estimate the current pose of the ego vehicle through onboard localization~\cite{hess2016real} in GPS-denied indoor environments. The Xavier NX then computes an ego-centric coarse topology map~\cite{amini2019variational}, which includes a highlighted routed map alongside an unrouted map, based on the ego vehicle's pose and a globally routed path. This path is planned using the Dijkstra algorithm, utilizing a sparse topological roadmap of the indoor corridor environments. The Arduino Nano receives commands from either the Xavier NX or a human driver, converting them into PWM signals for the steering and speed control motors of the platform.

\begin{figure}[h]
\includegraphics[width=0.48\textwidth]{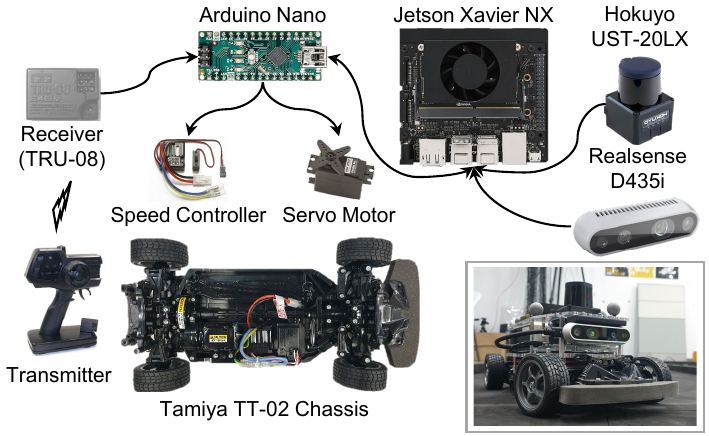}
\caption{Hardware system setup.
}
\label{fig:hardware}
\end{figure}

\subsection{Data Collection and Processing}
\label{appendix:data}
While collecting data, we record camera images, topology maps, and corresponding command signals for steering and throttle control from the human driver. These control signals are normalized to a range of [-1, +1]. We collect data for a total of 2 hours, amounting to 88,326 pairs of sensory input and labels in the environments of Env. 1 and Env. 2. The data is split into training and validation sets at a ratio of $80:20$. The camera image is cropped to a size of $440\times240$ pixels and then resized to $224\times224$ pixels for use in our network. For data augmentation, we apply random image shifts and corresponding steering angle adjustments, as outlined in~\cite{bojarski2016end}.

\begin{figure*}[t!]
\centering
\includegraphics[width=0.99\textwidth]{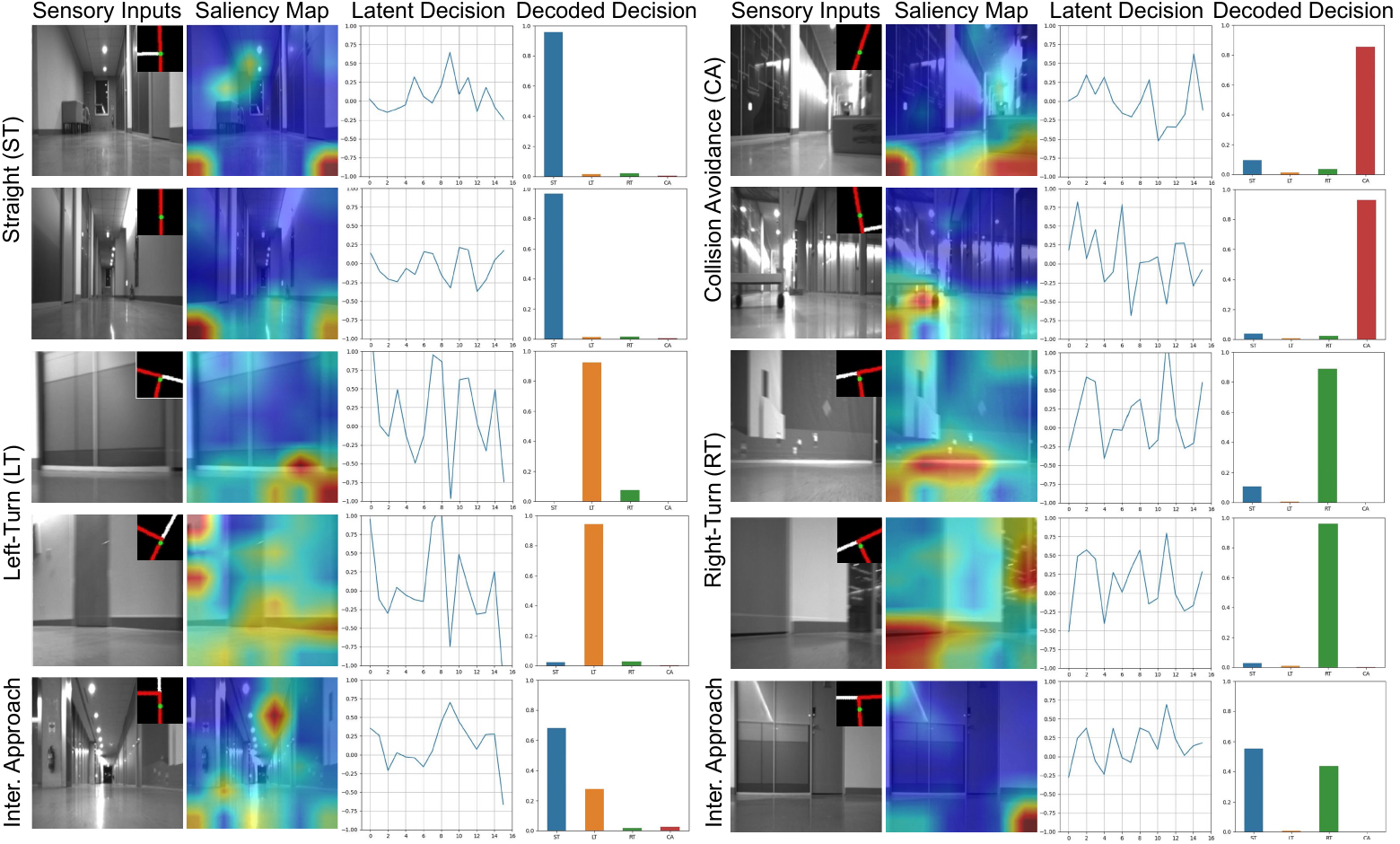}
\caption{Results of the spatial saliency maps, latent decisions, and decoded interpretable decisions, corresponding to given sensory inputs from various driving tasks. Latent decisions are plotted to represent their distribution, while the decoded decisions are presented as posterior probabilities between ST, LT, RT, and CA. 
}
\label{fig:interpretation}
\vspace{-15pt}
\end{figure*}
\subsection{Perceptual and Behavioral Interpretability}
\label{appendix:decoding}
Fig.~\ref{fig:interpretation} illustrates the results of perceptual and behavioral interpretation among various driving tasks. 
Using the spatial saliency map, we can explicitly interpret where the network focuses during autonomous navigation in real-world indoor environments. While the network does not specifically focus on any areas when driving straight through corridors, it shows strong spatial attention on the boundaries of intersections during turns, areas crucial for navigating the desired route. Similarly, upon encountering obstacles, the network generates spatial attention on the obstacle regions, further emphasizing critical areas for avoiding collisions. These results show that our network effectively identifies the regions with spatial importance in the visual sensory input during end-to-end autonomous driving, offering human engineers understandable insights into its perceptual processes.

Our model can also provide explainable decisions while performing end-to-end sensorimotor processing by decoding task-specific top-down latent decisions.
In the experiments, our method yields explainable decoded decisions, which are validly recognized as corresponding to the driving situation based on the sensory inputs (Fig.~\ref{fig:interpretation}). Even with varying environmental conditions in the driving scene, the top-down latent decision produces a similar distribution of neural values when the task-level context is analogous, resulting in accurate interpretations of behavioral intents.

Moreover, our method demonstrates both flexibility and scalability in interpretability. Drawing on previous quantitative results, we have consolidated straight driving tasks (ST, SI) into a single category, ST, by reconfiguring samples for the refitting of the SVMs. This underscores our method's ability to tailor the interpretation method to meet the specific needs of human engineers without having to retrain the original end-to-end network. Additionally, during navigation, we observe transition zones when the robot approaches intersections (\textit{Inter. Approach}), shifting from straight driving (ST) to turning (LT/RT). This transition presents a unique pattern, with both ST and LT/RT exhibiting high posterior probabilities simultaneously. Standing apart from the five predefined tasks (ST, SI, LT, RT, CA), this pattern suggests our method's capacity to uncover new behavioral tasks not previously identified by human engineers. As mentioned, SVM samples can be restructured if necessary to facilitate interpretation of these newly identified tasks.